%% file: main.tex
\documentclass[10pt,twocolumn,letterpaper]{article}

\usepackage[pagenumbers]{wacv} 


\definecolor{wacvblue}{rgb}{0.21,0.49,0.74}
\usepackage[pagebackref,breaklinks,colorlinks,allcolors=wacvblue]{hyperref}
\usepackage{graphicx}
\usepackage{amsmath}
\usepackage{booktabs}

\usepackage{graphicx}
\usepackage{multicol}
\usepackage{multirow}
\usepackage{makecell}
\usepackage{amssymb, pifont}
\definecolor{lightbluepub}{HTML}{D0E7FA}
\usepackage{afterpage}
\usepackage{float}
\usepackage{tikz}
\usetikzlibrary{decorations.pathreplacing}
\usepackage[accsupp]{axessibility}  

\def\wacvPaperID{56} 
\def\confName{WACV}
\def\confYear{2026}

\title{\texttt{SGPMIL}: Sparse Gaussian Process Multiple Instance Learning}

\author{
Andreas Lolos$^{1,3}$\thanks{Corresponding author: \texttt{andreaslolos@phys.uoa.gr}} \quad
Stergios Christodoulidis$^{4}$ \quad
Aris~L.~Moustakas$^{1,3}$ \quad
Jose Dolz$^{2}$ \\
Maria Vakalopoulou$^{3,4}$ \\
$^1$National and Kapodistrian University of Athens, Greece \quad
$^2$ÉTS Montréal, Canada \\
$^3$Archimedes, Athena Research Center, Greece \quad
$^4$CentraleSupélec, Université Paris-Saclay, France
}

\begin{document}
\maketitle

\input{sections/0_abstract}
\label{sec:abstract}

\section{Introduction}
\label{sec:intro}
\input{sections/1_intro}

\section{Related Work}
\label{sec:related}
\input{sections/2_related}

\section{Methodology}
\label{sec:methodology}
\input{sections/3_method}

\section{Experiments}
\label{sec:experiments}
\input{sections/4_experiments}

\section{Results \& Discussion}
\label{sec:results}
\input{sections/5_results}

\section{Conclusion}
\label{sec:conclusion}
\input{sections/6_conclusions}

\label{sec:aknowledgements}
\input{sections/8_aknowledgements}

{\small
\bibliographystyle{ieeenat_fullname}
\bibliography{main}
}

\end{document}


\maketitle

\section{Additional Experimental Results}


\begin{table}[htbp]
\centering
\begin{tabular}{lccc}
\toprule
 & ACC & AUC & ACE \\
\midrule
ABMIL & $.964^{\star}_{.010}$ & $\mathbf{.990}_{.005}$ & $.032^{\star}_{.011}$ \\
CLAM & $\underline{.978}_{.007}$ & $\underline{.986}_{.007}$ & $\underline{.021}_{.007}$ \\
TransMIL & $.962^{\star}_{.009}$ & $.980^{\star}_{.004}$ & $.029^{\star}_{.014}$ \\
DGRMIL & $.960^{\star}_{.012}$ & $.980^{\star}_{.010}$ & $.045^{\star}_{.016}$ \\
BayesMIL & $.976_{.007}$ & $.981^{\star}_{.006}$ & $\mathbf{.020}_{.007}$ \\
MixMIL & $.960^{\star}_{.009}$ & $.978^{\star}_{.007}$ & $.430^{\star}_{.002}$ \\
AGP & $.883^{\star}_{.021}$ & $.954^{\star}_{.019}$ & $.069^{\star}_{.013}$ \\
\textbf{SGPMIL} & $\mathbf{.980}_{.007}$ & $\underline{.986}_{.005}$ & $\underline{.021}_{.005}$ \\
\bottomrule
\end{tabular}
\caption{Slide-level performance on CAMELYON16. ${}^\star$ indicates significance based on a one-sided paired t-test across folds ($p < .05$).}
\label{tab:slide_level_CAMELYON16}
\end{table}

\begin{table}[htbp]
\centering
\begin{tabular}{lccc}
\toprule
 & ACC ($p$) & AUC ($p$) & ACE ($p$) \\
\midrule
ABMIL      & $\mathbf{1.19e{-3}}$ & $4.68\mathrm{e}{-1}$ & $\mathbf{2.72e{-3}}$ \\
CLAM       & $8.59\mathrm{e}{-2}$ & $2.40\mathrm{e}{-1}$ & $2.60\mathrm{e}{-1}$ \\
TransMIL   & $\mathbf{6.40e{-5}}$ & $\mathbf{5.66e{-3}}$ & $\mathbf{2.11e{-2}}$ \\
DGRMIL     & $\mathbf{3.15e{-4}}$ & $\mathbf{9.82e{-3}}$ & $\mathbf{1.17e{-3}}$ \\
BayesMIL   & $\mathbf{4.43e{-2}}$ & $\mathbf{3.07e{-5}}$ & $3.79\mathrm{e}{-1}$ \\
MixMIL     & $\mathbf{4.12e{-5}}$ & $\mathbf{4.28e{-3}}$ & $\mathbf{1.23e{-18}}$ \\
AGP        & $\mathbf{7.56e{-8}}$ & $\mathbf{1.09e{-4}}$ & $\mathbf{2.91e{-7}}$ \\
\bottomrule
\end{tabular}
\caption{$p$-values from one-sided paired t-tests comparing SGPMIL with baseline models across folds on CAMELYON16 (slide level). Bold values indicate $p < .05$.}
\label{tab:slide_level_CAMELYON16_pvalues}
\end{table}


\begin{table}[htbp]
\centering
\begin{tabular}{lccc}
\toprule
 & ACC & AUC & ACE \\
\midrule
ABMIL & $.953_{.003}$ & $.973_{.009}$ & $.039_{.008}$ \\
CLAM & $.934^{\star}_{.014}$ & $.953^{\star}_{.004}$ & $.056_{.016}$ \\
TransMIL & $.950_{.017}$ & $.970_{.012}$ & $.046_{.019}$ \\
DGRMIL & $.947_{.024}$ & $\underline{.974}_{.011}$ & $\underline{.038}_{.022}$ \\
BayesMIL & $\underline{.953}_{.023}$ & $.973_{.021}$ & $\mathbf{.033}_{.017}$ \\
MixMIL & $.925^{\star}_{.015}$ & $.963_{.014}$ & $.410^{\star}_{.025}$ \\
AGP & $.948^{\star}_{.026}$ & $\mathbf{.976}_{.014}$ & $.048_{.025}$ \\
\textbf{SGPMIL} & $\mathbf{.955}_{.021}$ & $.973_{.014}$ & $.047_{.027}$ \\
\bottomrule
\end{tabular}
\caption{Slide-level performance on TCGA-NSCLC. ${}^\star$ indicates significance based on a one-sided paired t-test across folds ($p < .05$).}
\label{tab:slide_level_TCGA}
\end{table}

\begin{table}[htbp]
\centering
\begin{tabular}{lccc}
\toprule
 & ACC ($p$) & AUC ($p$) & ACE ($p$) \\
\midrule
ABMIL      & $2.08\mathrm{e}{-1}$ & $3.14\mathrm{e}{-1}$ & $3.66\mathrm{e}{-1}$ \\
CLAM       & $\mathbf{3.93e{-2}}$ & $\mathbf{2.18e{-2}}$ & $1.50\mathrm{e}{-1}$ \\
TransMIL   & $\mathrm{5.15e{-2}}$ & $1.26\mathrm{e}{-1}$ & $2.83\mathrm{e}{-1}$ \\
DGRMIL     & $6.06\mathrm{e}{-2}$ & $2.59\mathrm{e}{-1}$ & $3.54\mathrm{e}{-1}$ \\
BayesMIL   & $1.61\mathrm{e}{-1}$ & $2.89\mathrm{e}{-1}$ & $4.71\mathrm{e}{-1}$ \\
MixMIL     & $\mathbf{1.19e{-2}}$ & $5.38\mathrm{e}{-2}$ & $\mathbf{1.32e{-4}}$ \\
AGP        & $\mathbf{4.40e{-2}}$ & $4.97\mathrm{e}{-1}$ & $2.22\mathrm{e}{-1}$ \\
\bottomrule
\end{tabular}
\caption{$p$-values from one-sided paired t-tests comparing SGPMIL with baseline models across folds on TCGA-NSCLC (slide level). Bold values indicate $p < .05$.}
\label{tab:slide_level_NSCLC_pvalues}
\end{table}


\begin{table}[htbp]
\centering
\begin{tabular}{lccc}
\toprule
 & ACC & $\kappa$ & ACE \\
\midrule
ABMIL & $.834^{\star}_{.064}$ & $.910^{\star}_{.028}$ & $.044^{\star}_{.015}$ \\
CLAM & $\underline{.867}^{\star}_{.061}$ & $.927^{\star}_{.025}$ & $.031^{\star}_{.018}$ \\
TransMIL & $.827^{\star}_{.074}$ & $.911^{\star}_{.030}$ & $.043^{\star}_{.021}$ \\
DGRMIL & $.843^{\star}_{.097}$ & $\underline{.933}^{\star}_{.047}$ & $.036^{\star}_{.025}$ \\
BayesMIL & $.850^{\star}_{.060}$ & $.926^{\star}_{.031}$ & $.031^{\star}_{.016}$ \\
MixMIL & $.690^{\star}_{.054}$ & $.870^{\star}_{.028}$ & $.180^{\star}_{.010}$ \\
AGP & $.802^{\star}_{.086}$ & $.906^{\star}_{.047}$ & $\mathbf{.026}_{.013}$ \\
\textbf{SGPMIL} & $\mathbf{.900}_{.065}$ & $\mathbf{.955}_{.037}$ & $\underline{.028}_{.022}$ \\
\bottomrule
\end{tabular}
\caption{Slide-level performance on PANDA. ${}^\star$ indicates significance based on a one-sided paired t-test across folds ($p < .05$).}
\label{tab:slide_level_PANDA}
\end{table}

\begin{table}[htbp]
\centering
\begin{tabular}{lccc}
\toprule
 & ACC ($p$) & $\kappa$ ($p$) & ACE ($p$) \\
\midrule
ABMIL      & $\mathbf{1.86e{-3}}$ & $\mathbf{8.04e{-4}}$ & $\mathbf{1.01e{-3}}$ \\
CLAM       & $\mathbf{7.51e{-3}}$ & $\mathbf{2.03e{-3}}$ & $\mathbf{4.44e{-3}}$ \\
TransMIL   & $\mathbf{4.50e{-3}}$ & $\mathbf{1.53e{-3}}$ & $\mathbf{1.56e{-3}}$ \\
DGRMIL     & $\mathbf{1.51e{-2}}$ & $\mathbf{3.50e{-2}}$ & $\mathbf{2.69e{-2}}$ \\
BayesMIL   & $\mathbf{1.60e{-3}}$ & $\mathbf{2.92e{-3}}$ & $\mathbf{6.92e{-3}}$ \\
MixMIL     & $\mathbf{2.23e{-3}}$ & $\mathbf{2.95e{-3}}$ & $\mathbf{8.19e{-5}}$ \\
AGP        & $\mathbf{1.57e{-2}}$ & $\mathbf{1.71e{-2}}$ & $2.31\mathrm{e}{-1}$ \\
\bottomrule
\end{tabular}
\caption{$p$-values from one-sided paired t-tests comparing SGPMIL with baseline models across folds on PANDA (slide level). Bold values indicate $p < .05$.}
\label{tab:slide_level_PANDA_pvalues}
\end{table}


\begin{table}[htbp]
\centering
\begin{tabular}{lccc}
\toprule
 & ACC & AUC & ACE \\
\midrule
ABMIL & $.694^{\star}_{.010}$ & $.852^{\star}_{.025}$ & $.175^{\star}_{.007}$ \\
CLAM & $.699^{\star}_{.034}$ & $.850^{\star}_{.021}$ & $.183^{\star}_{.011}$ \\
TransMIL & $.676^{\star}_{.005}$ & $.826^{\star}_{.032}$ & $.186^{\star}_{.012}$ \\
DGRMIL & $\underline{.703}^{\star}_{.033}$ & $.818^{\star}_{.035}$ & $.186^{\star}_{.023}$ \\
BayesMIL & $.648^{\star}_{.058}$ & $.829^{\star}_{.022}$ & $.183^{\star}_{.028}$ \\
MixMIL & $.662^{\star}_{.031}$ & $\underline{.855}^{\star}_{.010}$ & $.254^{\star}_{.002}$ \\
AGP & $.634^{\star}_{.030}$ & $.830^{\star}_{.010}$ & $\mathbf{.134}_{.014}$ \\
\textbf{SGPMIL} & $\mathbf{.736}_{.029}$ & $\mathbf{.870}_{.026}$ & $\underline{.142}_{.032}$ \\
\bottomrule
\end{tabular}
\caption{Slide-level performance on BRACS. ${}^\star$ indicates significance based on a one-sided paired t-test across folds ($p < .05$).}
\label{tab:slide_level_BRACS}
\end{table}

\begin{table}[htbp]
\centering
\begin{tabular}{lccc}
\toprule
 & ACC ($p$) & AUC ($p$) & ACE ($p$) \\
\midrule
ABMIL      & $\mathbf{1.81e{-3}}$ & $\mathbf{3.75e{-2}}$ & $\mathbf{1.53e{-2}}$ \\
CLAM       & $\mathbf{3.11e{-2}}$ & $\mathbf{4.64e{-2}}$ & $\mathbf{7.92e{-3}}$ \\
TransMIL   & $\mathbf{7.29e{-4}}$ & $\mathbf{1.30e{-2}}$ & $\mathbf{8.27e{-3}}$ \\
DGRMIL     & $\mathbf{3.07e{-2}}$ & $\mathbf{8.13e{-3}}$ & $\mathbf{6.68e{-3}}$ \\
BayesMIL   & $\mathbf{3.69e{-3}}$ & $\mathbf{8.01e{-3}}$ & $\mathbf{1.03e{-2}}$ \\
MixMIL     & $\mathbf{8.88e{-4}}$ & $\mathbf{2.94e{-2}}$ & $\mathbf{1.49e{-4}}$ \\
AGP        & $\mathbf{4.81e{-4}}$ & $\mathbf{3.85e{-3}}$ & $1.80\mathrm{e}{-1}$ \\
\bottomrule
\end{tabular}
\caption{$p$-values from one-sided paired t-tests comparing SGPMIL with baseline models across folds on BRACS (slide level). Bold values indicate $p < .05$.}
\label{tab:slide_level_BRACS_pvalues}
\end{table}


\begin{table}[htbp]
\centering
\begin{tabular}{lccc}
\toprule
 & ACC ($p$) & AUC ($p$) & ACE ($p$) \\
\midrule
ABMIL      & $\mathbf{1.27e{-4}}$ & $\mathbf{3.84e{-3}}$ & $\mathbf{6.41e{-4}}$ \\
CLAM       & $\mathbf{8.92e{-4}}$ & $\mathbf{2.29e{-3}}$ & $\mathbf{1.60e{-4}}$ \\
TransMIL   & $\mathbf{4.53e{-5}}$ & $\mathbf{1.14e{-3}}$ & $\mathbf{4.08e{-2}}$ \\
DGRMIL     & $\mathbf{8.89e{-3}}$ & $\mathbf{4.82e{-2}}$ & $\mathbf{6.19e{-4}}$ \\
BayesMIL   & $\mathbf{2.31e{-4}}$ & $\mathbf{1.12e{-4}}$ & $\mathbf{1.51e{-5}}$ \\
MixMIL     & $\mathbf{3.03e{-5}}$ & $\mathbf{3.49e{-3}}$ & $\mathbf{3.96e{-7}}$ \\
AGP        & $\mathbf{4.15e{-3}}$ & $\mathbf{6.15e{-3}}$ & $9.75\mathrm{e}{-1}$ \\
\bottomrule
\end{tabular}
\caption{$p$-values from one-sided paired t-tests comparing SGPMIL with baseline models across folds on CAMELYON16 (instance level). Bold values indicate $p < .05$.}
\label{tab:instance_level_CAMELYON16_pvalues}
\end{table}

\begin{table}[htbp]
\centering
\begin{tabular}{lccc}
\toprule
 & ACC ($p$) & AUC ($p$) & ACE ($p$) \\
\midrule
ABMIL      & $\mathbf{1.03e{-5}}$ & $\mathbf{2.26e{-5}}$ & $8.75\mathrm{e}{-2}$ \\
CLAM       & $\mathbf{1.50e{-5}}$ & $\mathbf{2.17e{-4}}$ & $7.37\mathrm{e}{-2}$ \\
TransMIL   & $\mathbf{3.75e{-3}}$ & $\mathbf{6.65e{-3}}$ & $\mathbf{4.17e{-3}}$ \\
DGRMIL     & $\mathbf{1.79e{-4}}$ & $\mathbf{1.20e{-4}}$ & $\mathbf{8.47e{-3}}$ \\
BayesMIL   & $\mathbf{9.11e{-5}}$ & $\mathbf{3.22e{-4}}$ & $\mathbf{5.40e{-3}}$ \\
MixMIL     & $\mathbf{1.01e{-5}}$ & $\mathbf{2.81e{-6}}$ & $9.11\mathrm{e}{-1}$ \\
AGP        & $\mathbf{3.17e{-2}}$ & $\mathbf{3.17e{-3}}$ & $\mathbf{8.10e{-7}}$ \\
\bottomrule
\end{tabular}
\caption{$p$-values from one-sided paired t-tests comparing SGPMIL with baseline models across folds on BRACS (instance level). Bold values indicate $p < 0.05$.}
\label{tab:instance_level_BRACS_pvalues}
\end{table}


\begin{table}[htbp]
\centering
    \centering
    \label{tab:inducing_points_ablations_sgpmil}
    \begin{tabular}{lcccccc}
        \toprule
        & \multicolumn{3}{c}{WSI} & \multicolumn{3}{c}{INSTANCE} \\
        \cmidrule(lr){2-4} \cmidrule(lr){5-7}
        $U$ & ACC & AUC & ACE & ACC & AUC & ACE \\
        \midrule
        16 & .971 & .977 & .045 & \underline{.893} & .963 & .271 \\
        32 & \underline{.979} & \underline{.985} & \underline{.026} & .880 & \underline{.970} & \textbf{.046} \\
        64 & .964 & .981 & .041 & .832 & .910 & .210 \\
        80 & \textbf{.980} & \textbf{.986} & \textbf{.021} & \textbf{.914} & \textbf{.973} & \underline{.051} \\
        \bottomrule
    \end{tabular}
    \caption{Performance of SGPMIL on CAMELYON16 for varying inducing points.}
\end{table}
\hfill
\begin{table}[t]
    \centering
    \begin{tabular}{lccc}
    \toprule
    Model & Training (s) & Inference (s) & Params (M) \\
    \midrule
    ABMIL      & $5.5^{\star}$ & $0.8^{\star}$ & 0.66 \\
    CLAM       & $7.0$         & $0.9$         & 0.92 \\
    TransMIL   & $13.4$        & $1.2$         & 2.67 \\
    DGRMIL     & $16.7$        & $1.5$         & 4.34 \\
    BayesMIL   & $9.5$         & $1.1$         & 1.32 \\
    MixMIL     & $11$          & $1.5$         & 1.57 \\
    AGP        & $31.0$        & $6.0$         & 1.21 \\
    \textbf{SGPMIL} & $9.0^{\star\star}$ & $1.0^{\star\star}$ & 1.21 \\
    \bottomrule
    \end{tabular}
    \caption{Training and inference times (in seconds) and model sizes (number of trainable parameters in millions, M). 
    Training times are averaged over 30 epochs, while inference times correspond to processing the full test set of $129$ slides. For MixMIL, we report the full variational posterior variant (non-mean field approximation).
    ${}^{\star}$ marks the overall fastest model (ABMIL) and ${}^{\star\star}$ highlights the fastest probabilistic model among AGP, BayesMIL, MixMIL and SGPMIL.}
    \label{tab:timing_comparison}
\end{table}


\begin{figure*}[htbp]
    \centering
    \includegraphics[width=0.98\linewidth]{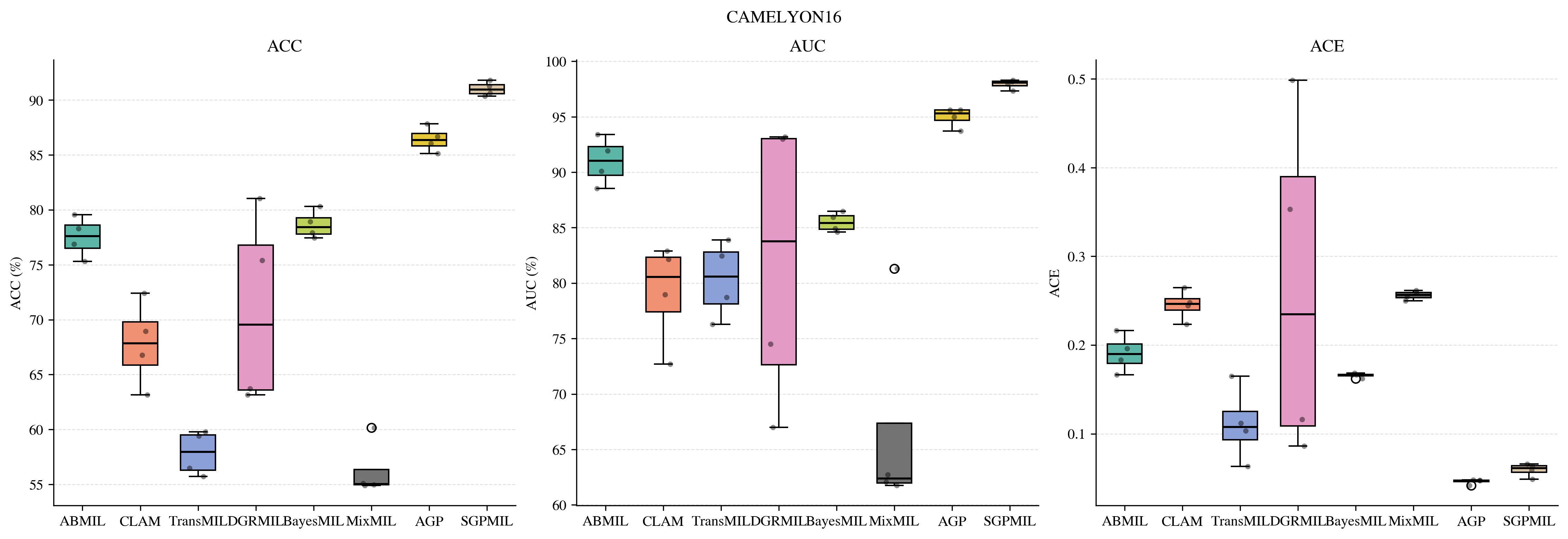}
    \caption{Instance-level performance across folds on CAMELYON16. Boxplots show the distribution of accuracy (ACC), area under the ROC curve (AUC), and adaptive expected calibration error (ACE) for each model. ${}^\star$ denotes significance according to one-sided paired t-tests ($p < .05$).}

\label{fig:supplementary_heatmaps}
\end{figure*}

\begin{figure*}[htbp]
    \centering
    \includegraphics[width=0.98\linewidth]{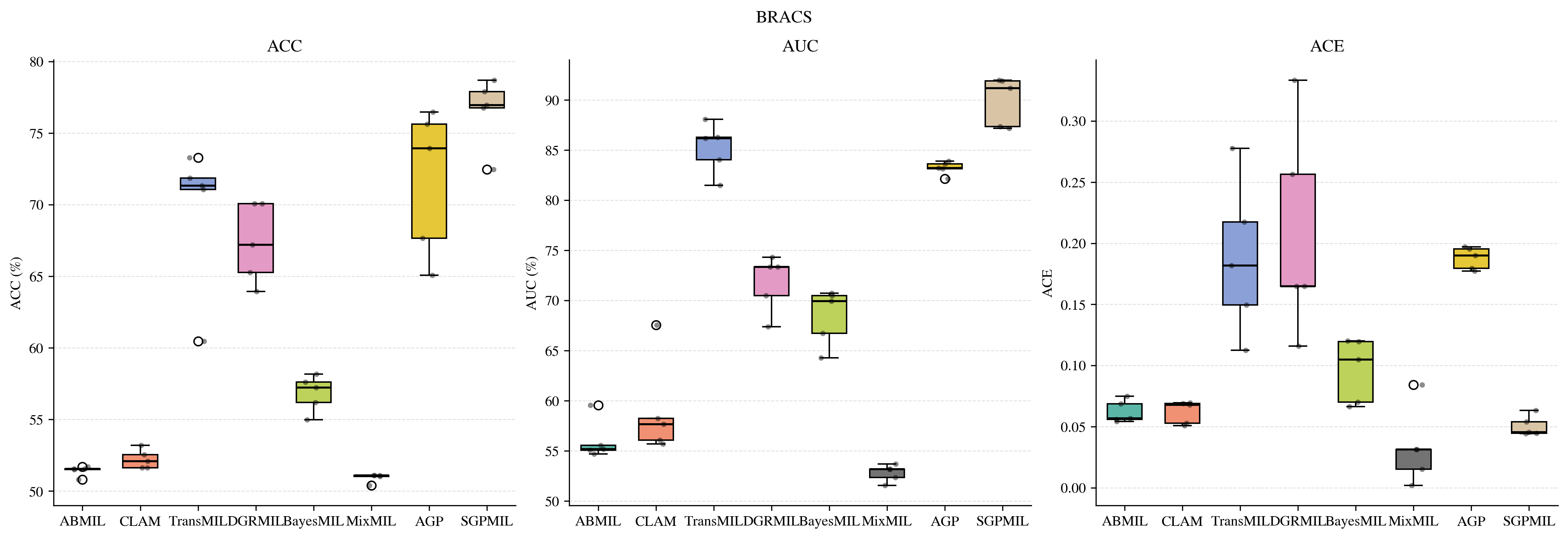}
    \caption{Instance-level performance across folds on BRACS. Boxplots show the distribution of accuracy (ACC), area under the ROC curve (AUC), and adaptive expected calibration error (ACE) for each model. ${}^\star$ denotes significance according to one-sided paired t-tests ($p < .05$).}

\label{fig:supplementary_heatmaps}
\end{figure*}


\newpage

\begin{figure*}[htbp]
    \centering
    \includegraphics[width=0.98\linewidth]{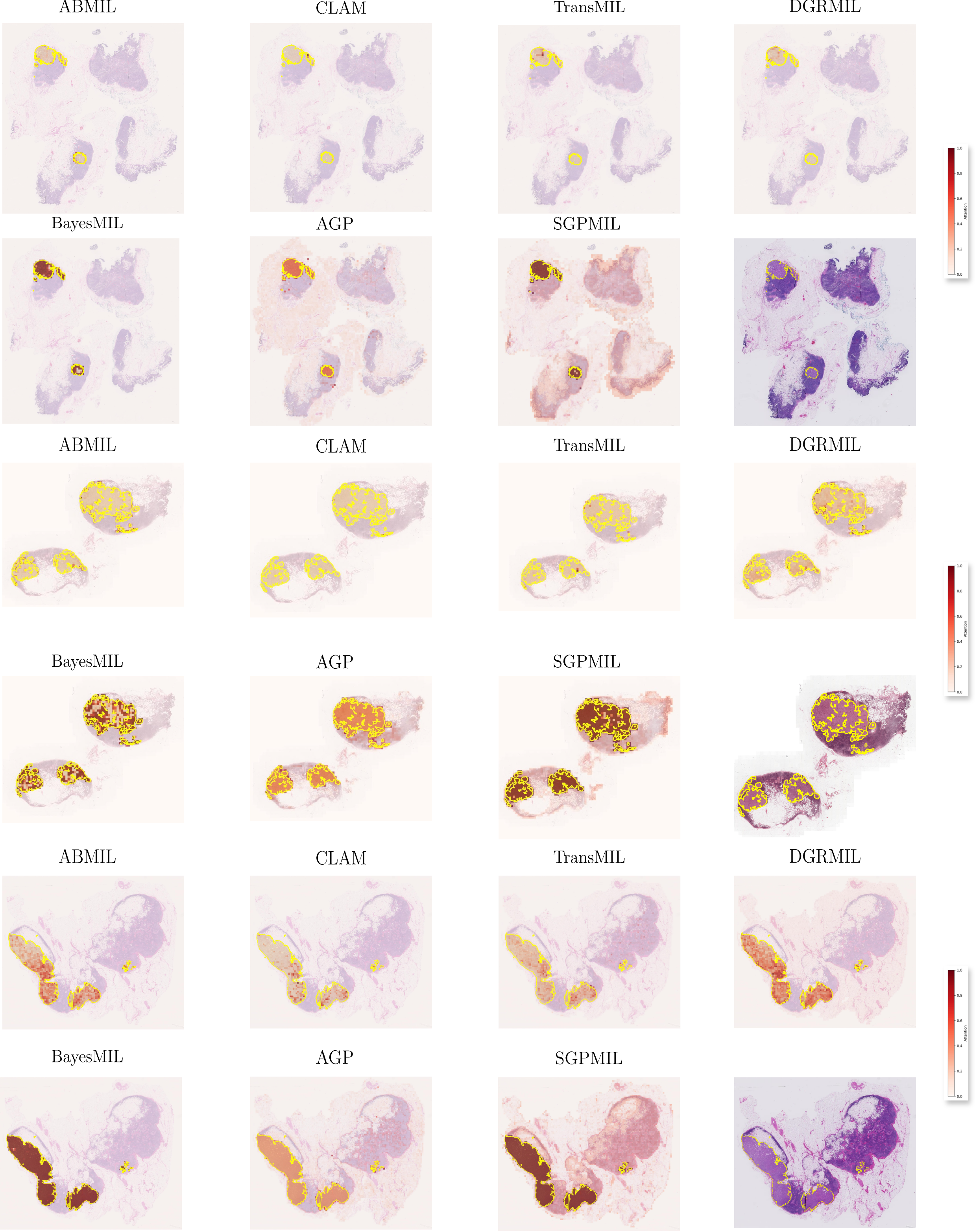}
    \caption{Additional attention heatmaps on CAMELYON16. Ground-truth tumor annotations are overlaid as yellow contours.}
\label{fig:supplementary_heatmaps}
\end{figure*}

\newpage

\begin{figure*}[htbp]
    \centering
    \includegraphics[width=0.98\linewidth]{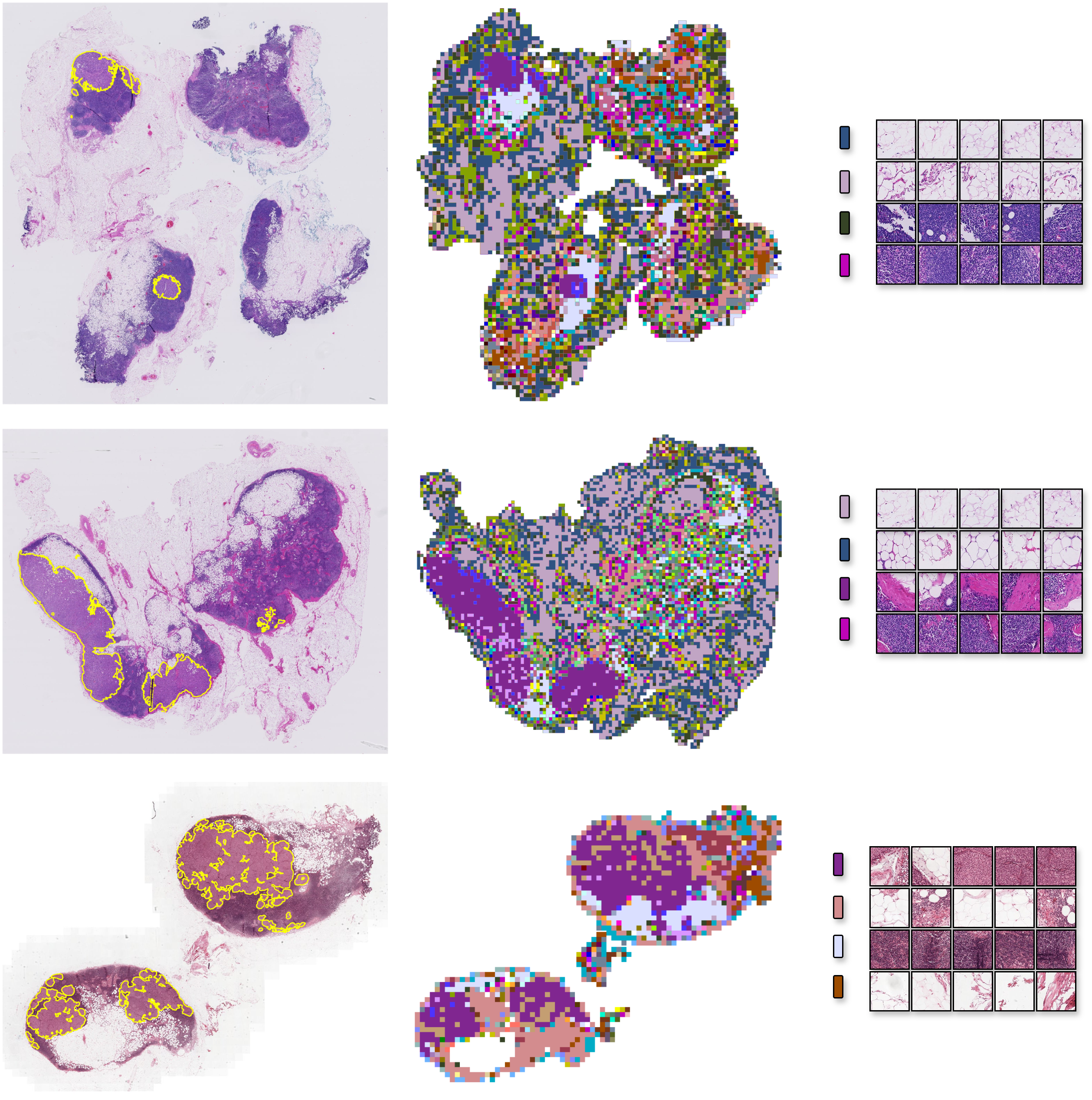}
\caption{Additional inducing point label maps. Left: CAMELYON16 test set WSIs with ground-truth tumor annotations (yellow contours). Middle: label maps where each patch embedding is assigned to its most similar inducing point based on cosine similarity. Right: top-5 most similar patches for selected inducing points. These visualizations highlight how inducing points specialize in different tissue morphologies, including tumor, stroma, and interface regions.}

\label{fig:supplementary_labelmaps}
\end{figure*}

\begin{table}[htbp]
\centering
\label{tab:kernel_ablations_sgpmil_wsi}
\begin{tabular}{lccc}
    \toprule
     & \multicolumn{3}{c}{WSI} \\
     \cmidrule{2-4}
    Kernel & ACC & AUC & ACE \\
    \midrule
    RBF & $.980_{.007}$ & $.986_{.005}$ & $.021_{.005}$ \\
    Linear & $.973_{.010}$ & $.985_{.005}$ & $.048_{.008}$ \\
    Matern & $.985_{.006}$ & $.988_{.003}$ & $.024_{.006}$ \\
    \bottomrule
\end{tabular}
\caption{Performance of SGPMIL on CAMELYON16 for various kernels, across folds.}
\end{table}

\begin{table}[htbp]
\centering
\label{tab:kernel_ablations_sgpmil_instance}
\begin{tabular}{lccc}
    \toprule
     & \multicolumn{3}{c}{INSTANCE} \\
     \cmidrule{2-4}
    Kernel & ACC & AUC & ACE \\
    \midrule
    RBF & $.914_{.006}$ & $.973_{.004}$ & $.051_{.008}$ \\
    Linear & $.851_{.017}$ & $.950_{.020}$ & $.113_{.016}$ \\
    Matern & $.850_{.011}$ & $.960_{.012}$ & $.062_{.014}$ \\
    \bottomrule
\end{tabular}
\caption{Performance of SGPMIL on CAMELYON16 for various kernels, across folds.}
\end{table}

\newpage

\begin{figure*}[htbp]
    \centering
    \includegraphics[width=0.9\linewidth]{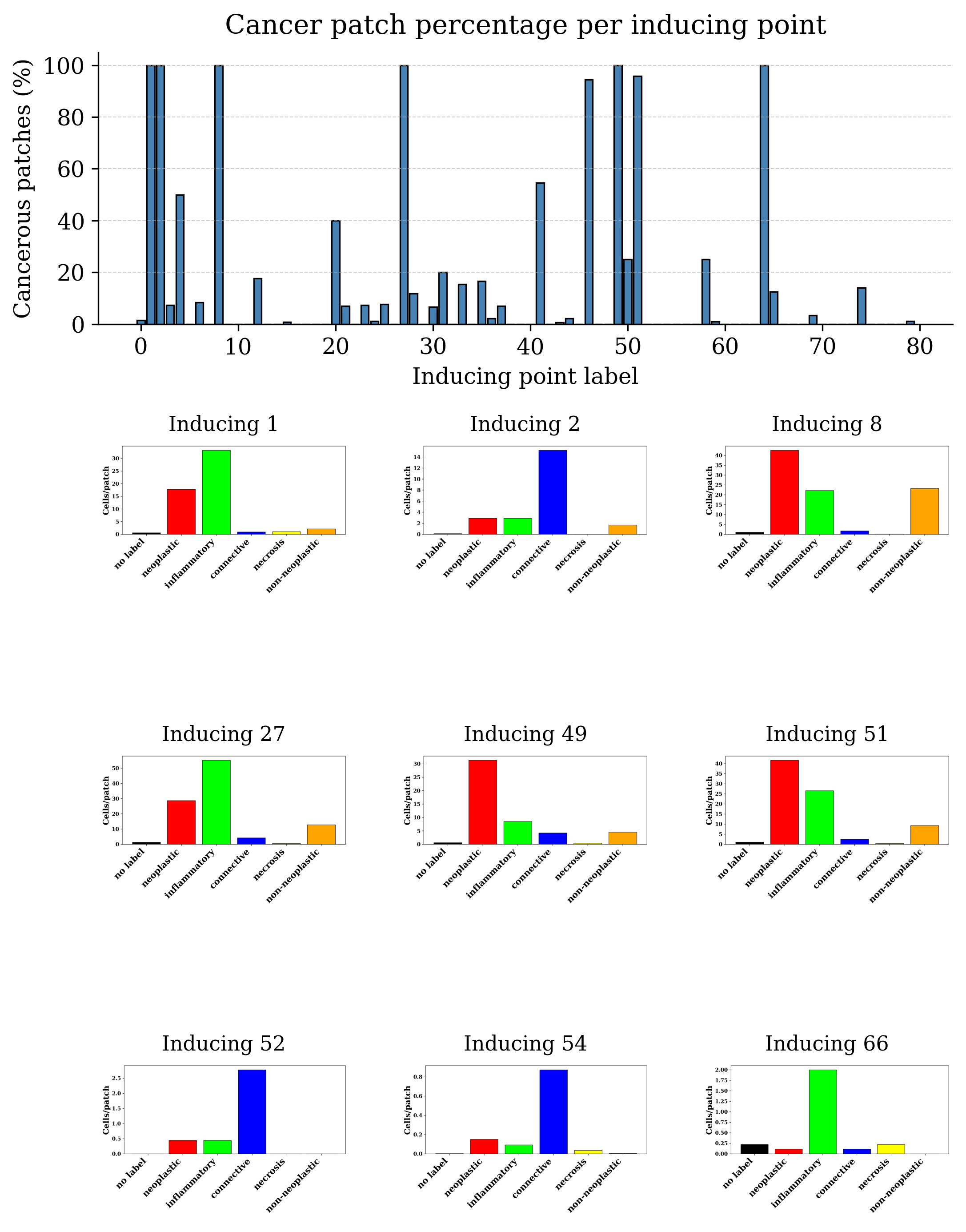}
    \caption{Cancer patch percentage per inducing point (top) and nucleus type distributions (bottom) for selected inducing points in CAMELYON16. Percentages were computed as the fraction of cancerous to total patches assigned to each inducing point in the slide. Inducing points 1, 2, 8, 27, 49, and 51 are highly enriched in cancerous regions and show a predominance of neoplastic nuclei, with inflammatory cells often the next most abundant type. By contrast, inducing points not associated with cancerous patches (52, 54, 66) mainly reflect connective or non-tumor tissue, with sparse isolated tumor cells, consistent with the lack of pixel-level annotations in CAMELYON16.}

\label{fig:supplementary_hovernet}
\end{figure*}

\begin{figure*}[htbp]
    \centering
    \includegraphics[width=\linewidth, height=0.9\textheight, keepaspectratio]{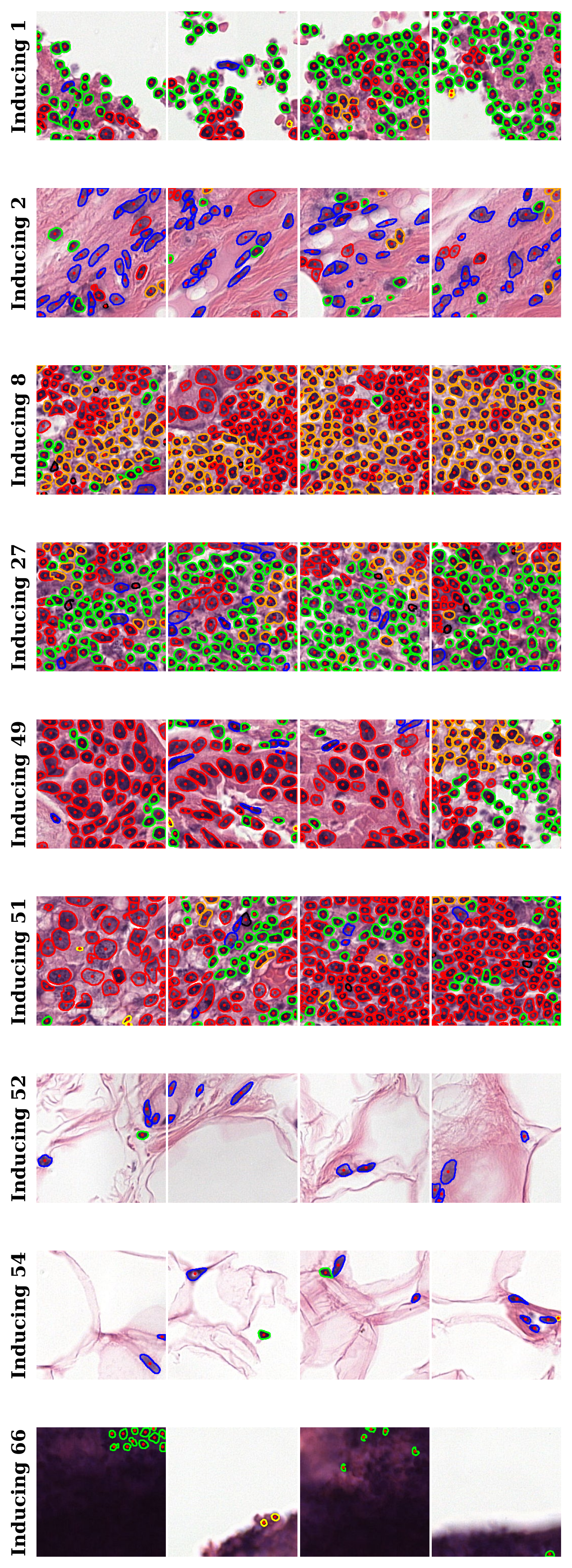}

    \caption{Representative image patches most similar to inducing points in CAMELYON16 slide 75. Each row corresponds to one inducing point (labels shown on the left), with four example patches illustrating the tissue context and associated nuclear types as segmented by HoverNet. Inducing points 1, 2, 8, 27, 49, and 51 predominantly align with tumor-rich regions, while others (e.g., 52, 54, 66) capture connective or non-tumor tissue, occasionally containing isolated tumor cells.}

\label{fig:supplementary_hovernet_cells}
\end{figure*}

\begin{table}[ht]
    \centering
    \begin{tabular}{@{\hskip 2pt}l@{\hskip 2pt}c@{\hskip 2pt}c@{\hskip 2pt}c@{\hskip 2pt}c@{\hskip 2pt}c@{\hskip 2pt}c}
    \toprule
     & \multicolumn{2}{c}{Bag-level}  & \multicolumn{4}{c}{Instance-level} \\    
    \cmidrule(lr){2-3} \cmidrule(lr){4-7}
    & AUC & ACE & F1 & FROC & AUC & ACE \\
    \midrule
    ABMIL     & $0.986$ & $0.030$ & $0.958$ & $0.816$ & $0.987$ & $0.033$ \\
    CLAM      & $0.989$ & $0.021$ & $0.952$ & $0.813$ & $\mathbf{0.999}$ & $0.039$ \\
    TransMIL  & $0.990$ & $0.021$ & $\mathbf{0.980}$ & $0.826$ & $0.993$ & $0.025$ \\
    DGRMIL    & $\mathbf{0.996}$ & $\underline{0.016}$ & $\underline{0.979}$ & $\underline{0.827}$ & $0.996$ & $\mathbf{0.013}$ \\
    BayesMIL  & $0.990$ & $0.019$ & $0.968$ & $0.824$ & $0.994$ & $\underline{0.019}$ \\
    AGP       & $0.994$ & $0.046$ & $0.957$ & $0.820$ & $0.981$ & $0.083$ \\
    SGPMIL      & $\underline{0.995}$ & $\mathbf{0.009}$ & $0.978$ & $\mathbf{0.830}$ & $\underline{0.998}$ & $0.050$ \\
    \bottomrule
    \end{tabular}
    \caption{Bag-level and instance-level performance for the MNIST-bags dataset. Bags are comprised of 9 instances each. All baselines are trained end-to-end with a CNN as initial feature extractor.}
    \label{tab:combined_bags_instance_mnist_results}
\end{table}

%% file: sections/0_abstract.tex
\begin{abstract}
Multiple Instance Learning (MIL) offers a natural solution for settings where only coarse, bag-level labels are available, without having access to instance-level annotations. This is usually the case in digital pathology, which consists of gigapixel-sized images. While deterministic attention-based MIL approaches achieve strong bag-level performance, they often overlook the uncertainty inherent in instance relevance. In this paper, we address the lack of uncertainty quantification in instance-level attention scores by introducing \textbf{SGPMIL}, a new probabilistic attention-based MIL framework grounded in Sparse Gaussian Processes (SGP). By learning a posterior distribution over attention scores, SGPMIL enables principled uncertainty estimation, resulting in more reliable and calibrated instance relevance maps. Our approach not only preserves competitive bag-level performance but also significantly improves the quality and interpretability of instance-level predictions under uncertainty. SGPMIL extends prior work by introducing feature scaling in the SGP predictive mean function, leading to faster training, improved efficiency, and enhanced instance-level performance. 
Extensive experiments on multiple well-established digital pathology datasets highlight the effectiveness of our approach across both bag- and instance-level evaluations. Our code is available at \url{https://github.com/mandlos/SGPMIL}.
\end{abstract}

%% file: sections/1_intro.tex
Analyzing whole slide images (WSIs) is a critical task in digital pathology, central to medical diagnostics and treatment planning. Due to their gigapixel scale, WSIs are prohibitively expensive and time-consuming to annotate at the pixel or region level. Multiple Instance Learning (MIL) offers a practical solution by enabling model training using only coarse, slide-level labels, bypassing the need for exhaustive annotations~\cite{dietterich1997solving, maron1997framework, quellec2017multiple, ilse2018attention}. In the MIL framework, each data point (a ``bag'') consists of many instances (e.g., image patches), but only the bag label is provided during training. This setup makes MIL especially well-suited for WSI classification, subtype identification, and survival prediction~\cite{ilse2018attention, lu2021clam, panther}.

Building on recent progress in MIL for WSI analysis, embedding-based MIL models with attention pooling have demonstrated strong performance at the bag-level while offering interpretable heatmaps that highlight task-relevant tissue regions~\cite{ilse2018attention, shao2021transmil, zhu2024dgrmil, yufei2022bayes}.
These methods rely on attention mechanisms to weigh instance contributions, yet the resulting scores are often interpreted heuristically as indicators of instance relevance. In this work, we introduce a probabilistic formulation of the attention mechanism by learning a posterior distribution over attention weights. This formulation enables uncertainty estimation and enhances the interpretability of instance-level predictions, while maintaining strong bag-level performance.

Recent attention-based MIL aggregation schemes have demonstrated impressive bag-level performance across various metrics and datasets \cite{ilse2018attention,lu2021clam,shao2021transmil, zhu2024dgrmil, mixmil}. These results have been further enhanced by leveraging recent foundation models, which provide powerful instance-level representations.
The aforementioned MIL works showcase attention heatmaps of test set cases for explainability purposes, where these heatmaps seem to focus on instances of interest which, for example, in a binary classification setting indicate the presence or absence of a positive class instance. While the attained bag-level performance is impressive, attention values are often interpreted as class-specific instance discriminators. We hypothesize that explicitly modeling uncertainty over these attention scores---by learning a posterior distribution---can improve instance-level performance, while preserving strong bag-level accuracy.

Probabilistic deep learning offers principled tools for uncertainty estimation and improved model calibration \cite{ghahramani2015probabilistic}, with methods such as Batch-Ensemble \cite{wen2020batchensemble,dusenberry2020efficient}, Concrete Dropout \cite{gal_concrete_dropout}, and multiplicative Gaussian noise \cite{kingma2015variationaldropoutlocalreparameterization, vdo_molchanov,cui2021bayesian} approximating posterior distributions through ensembles, dropout, or noise injection.
This is particularly valuable in MIL, where probabilistic frameworks can clarify prediction confidence, statistical properties, and the inner workings of pooling operations \cite{yufei2022bayes, haussmann2017variational, mixmil}. Prior works have explored uncertainty-driven MIL in various forms, including Bayesian CNNs under full supervision \cite{thiagarajan2021explanation}, Bayesian instance-based MIL under data scarcity and label noise \cite{haussmann2017variational, morales2024introducing}, uncertainty-driven aggregation models that refine attention or instance selection scores \cite{yufei2022bayes, schmidt2023probabilistic, udmil, mixmil, struski2023promil} and OOD uncertainty estimation \cite{linmans2023predictive}. Conformal prediction (CP) provides an alternative, distribution-free framework that yields uncertainty sets with marginal coverage guarantees, though it requires a separate calibration set and exchangeability assumptions~\cite{conformal1}. Furthermore, Bayesian extensions of the embedding-based MIL have explored the performance/explainability gains that one has by inducing uncertainty to the attention scoring mechanism and imposing prior regularization \cite{yufei2022bayes, schmidt2023probabilistic}. These approaches typically incorporate variational inference and Monte Carlo (MC) sampling to model distributional uncertainty at the instance and bag-level. Moreover, current challenges in probabilistic MIL approaches include numerical instability during training~\cite{schmidt2023probabilistic}, limited evaluation on multiple datasets and tasks, or the need for an \textit{a priori} decision on the range of instance spatial correlations in the scoring function~\cite{yufei2022bayes}.

In this paper, we introduce \textbf{SGPMIL}, an efficient and robust probabilistic attention-based MIL framework grounded in Sparse Gaussian Processes (SGP). By designing a novel probabilistic formulation, our main contributions are structured along the following axes:

\noindent \textbf{(i) A fast and efficient probabilistic framework.} SGPMIL builds upon prior work on SGP for MIL and introduces three key innovations: (1) a feature scaling term in the predictive mean of the learnable attention SGP variational posterior, which improves representational flexibility and adaptation to varying input scales; (2) a more stable and robust sampling strategy from the variational posterior during training, addressing numerical instabilities in Gaussian Process-based MIL approaches; and (3) an interpretable instance-level attention normalization mechanism across attention samples, enabling more consistent and meaningful instance-level predictions under the MIL setting.

\noindent  \textbf{(ii) Instance-level performance / Interpretability.} Our Sparse Gaussian Process formulation of the variational posterior, combined with the proposed across-sample attention normalization, enables the model to assign high attention values to task-relevant instances in an inherently unsupervised manner. This design not only adheres to the core MIL assumption—where only a subset of instances in a bag contribute to the label—but also enhances interpretability by producing consistent and informative attention maps that reflect the model’s confidence and inductive biases.

\noindent  \textbf{(iii) Prediction uncertainty estimation.} Our framework provides calibrated uncertainty estimates by leveraging sampling from the variational posterior, enabling principled uncertainty quantification at the slide level. We empirically demonstrate that predictive uncertainty correlates with classification correctness, highlighting its utility in identifying uncertain or failure-prone predictions. This makes the model particularly suitable for deployment in safety-critical scenarios, such as medical image processing.

Extensive experiments across multiple datasets demonstrate on-par bag-level performance on binary classification tasks, with \textbf{SGPMIL} outperforming leading MIL approaches on multiclass datasets. Furthermore, we provide both intuitive motivation and empirical validation for the hypothesis that the inducing point approach in our MIL formulation captures salient histological morphologies and effectively focuses on task-relevant instances, leading to superior instance-level performance.

%% file: sections/2_related.tex
\subsection{Deterministic frameworks}
Many embedding-based MIL approaches compute instance-level attention scores through fixed parametric transformations. Ilse \etal~\cite{ilse2018attention} introduce attention-based deep MIL (ABMIL), where the slide-level representation is computed as a weighted average of instance embeddings. The attention weights are derived via a gated mechanism, allowing the model to learn instance relevance in a differentiable, end-to-end manner. 
Clustering-constrained attention MIL (CLAM) \cite{lu2021clam} extended the gated attention mechanism by introducing an extra clustering objective via a smooth SVM loss. Li \etal presented Dual-Stream MIL Network (DSMIL)~\cite{li2021dsmil}, a dual-attention mechanism that combines instance and bag-level attention, where the most critical instance is selected via a classifier stream and its features guide the aggregation of the remaining patches. This is coupled with multiscale feature extraction and contrastive pretraining, leading to improved tumor detection performance. 
Transformer-based MIL (TransMIL) \cite{shao2021transmil} made use of two self-attention heads \cite{vaswani2023attentionneed} to weigh the instances and introduce spatial correlations using a novel Pyramid Position Encoding Generator (PPEG) achieving strong performance in bag-level classification metrics. Rymarczyk \etal \cite{rymarczyk2021kernel} replace dot-product attention with kernel functions, underscoring the usefulness of kernelized self-attention for modeling instance correlations. More recent works like DGRMIL \cite{zhu2024dgrmil} model instance correlation by introducing cross-attention across instances and a set of learnable tokens which are trained by a positive instance alignment mechanism and a novel diversity loss term.

Despite strong bag-level performance, existing attention-based MIL methods rarely assess how well attention mechanisms identify task-relevant instances. Being deterministic, they also lack the capacity to estimate prediction uncertainty—crucial in medical applications. We address these gaps by introducing a probabilistic attention mechanism and conducting systematic instance-level evaluation, demonstrating improved identification of task-relevant instances along with principled uncertainty estimation.

\subsection{Probabilistic frameworks} 
AGP \cite{schmidt2023probabilistic} integrates a SGP layer within a multi-layer perceptron (MLP) to model attention distributions over instances for classification tasks. Given input embeddings, the SGP layer computes a variational multivariate normal posterior distribution over the attention values associated with each instance in a bag; Monte Carlo (MC) sampling yields multiple attention values, which are used to reweigh the instance embeddings, resulting in multiple samples of bag-level representations that are subsequently classified. The model is trained by maximizing the Evidence Lower Bound (ELBO) with respect to the model parameters. Another related approach is BayesMIL \cite{yufei2022bayes}, which addresses the interpretability of attention mechanism in MIL by modeling probabilistic instance weights. Specifically, they introduce multiplicative Gaussian noise \cite{kingma2015variationaldropoutlocalreparameterization}, into the projection layers and the classifier. 
To capture spatial correlations among instance attention distributions, they incorporate a conditional random field (CRF) on top of the learned attention distributions. MixMIL \cite{mixmil} integrates generalized linear mixed models with attention-based MIL by placing Gaussian priors on both instance and bag-level effects, injecting uncertainty into attention scores and estimating respective posterior distributions through variational inference; while robust and interpretable, its largely linear formulation may hinder performance on complex datasets.

Despite their promise, existing probabilistic approaches lack evaluations with pathology foundation encoders on cancer datasets, extensive instance-level evaluation on other datasets or rely on fixed spatial smoothing kernels, further requiring manual tuning~\cite{yufei2022bayes}. AGP~\cite{schmidt2023probabilistic}, in particular, faces numerical instabilities and scalability issues on larger datasets, and does not assess instance-level performance. In this work, we address these limitations by introducing a feature scaling term in the mean of the SGP variational posterior distribution, relaxed attention normalization and an efficient sampling strategy during training in AGP, addressing different instabilities, while  conducting a comprehensive evaluation of both bag- and instance-level performance on the widely used public benchmarks.

\paragraph{Gaussian Processes}
Gaussian Processes (GPs) offer a powerful, flexible, and elegant framework for regression, as is indicated by an extensive line of works \cite{snelson2005sparse, kim2010gaussian, titsias2009variational}. As non-parametric standalone regressors, they mitigate the need for application-specific model architectures and \textit{a priori} decision on model complexity \cite{jakkala2021deep, rasmussen2006gaussian}. Furthermore, \cite{neal2012bayesian, lee2018deepneuralnetworksgaussian} showcase the equivalence of GPs to an infinite-width single-layer fully-connected neural network with an i.i.d. prior over its parameters and with infinitely wide deep networks respectively which in turn are known to be universal function approximators \cite{hornik1989multilayer}. Yet, these benefits come at the cost of time and storage complexity of the order of $\mathcal{O}(n^3)$ and $\mathcal{O}(n^2)$ respectively mainly due to inverting an $n\times n$ kernel matrix, $n$ being the number of training points, which is prohibitive for large datasets \cite{murphy2023probabilistic}. To overcome the time and storage cost bottleneck, Sparse Gaussian Processes were originally introduced by \cite{snelson2005sparse} and subsequently extended to allow for mini-batch training \cite{hensman2015scalable}. We leverage the scalable variational GPs framework introduced in \cite{hensman2015scalable} to induce uncertainty over instance attention distributions, and incorporate distributional correlation through kernel-based terms, showing that modeling such semantic correlations substantially improves instance-level performance.

%% file: sections/3_method.tex
\subsection{Preliminaries}
\begin{figure*}[h!]
    \centering
    \includegraphics[width=0.92\linewidth]{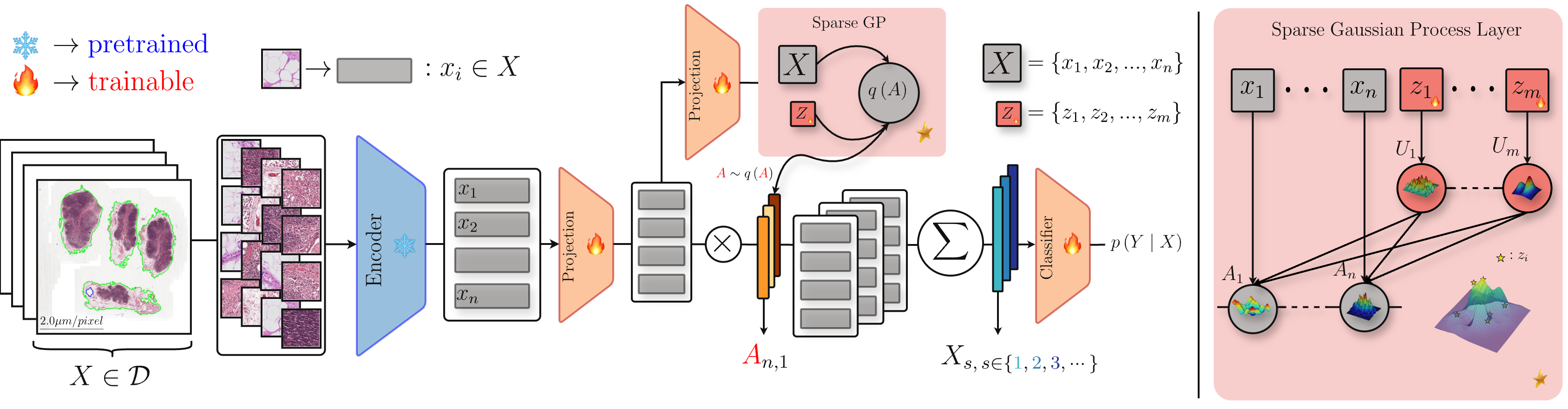}
    \vspace{-3mm}
    \caption{SGPMIL architecture overview. On the left, we illustrate the complete processing pipeline: WSIs are segmented and tiled into patches, which are then encoded using a frozen foundation model. The resulting patch embeddings are passed through an attention-based MIL head incorporating a probabilistic SGP mechanism. The attention-weighted embeddings are summed and projected through a trainable MLP. On the right, we highlight the probabilistic attention component. The SGP layer receives the embeddings along with learnable inducing points and infers a variational posterior over patch-level attention scores. Multiple attention samples are drawn to reweigh the embeddings, which are then aggregated into stochastic slide-level and classified via a linear layer followed by softmax.}

\label{fig:NNGP_pipeline_figure}
\end{figure*}

\noindent \textbf{Multiple Instance Learning (MIL)} In MIL, the input is a bag of instances, $X_i = \{x_{1}, \ldots, x_{K_i}\}$ with k-th instance $x_{k} \in \mathbb{R}^D$, associated with a bag label $Y_i \in \mathbb{N}$, while the instance labels $\{y_{1}, \ldots, y_{K_i}\}$ remain unobserved during training; here $K_i$ denotes the number of instances in bag $i$. A dataset consists of $N$ bag-label pairs, denoted as $\mathcal{D} = {(X_i, Y_i)}_{i=1}^N$, and let $\mathcal{C}=\{1,\ldots,C\}$ denote the set of class labels. The standard MIL assumption is defined as:
\begin{equation}
    Y_i = 
    \begin{cases} 
        0, & \text{if } \sum_{k=1}^{K_i} y_{k} = 0 \\[5pt]
        c, & \text{if } \exists\, k \text{ such that } y_{k} = c,c\in \mathcal{C}
    \end{cases}
\end{equation}

\noindent \textbf{Sparse Gaussian Processes (SGPs)} Following standard notation, given a training dataset $\{\left(X_i, Y_i\right)\}_{i=1}^{n}$ of size $n$, where $Y_i=F_i+\epsilon_i$ is a noisy observation of the noise-free real-valued latent function $F(\cdot):X\in\mathcal{X}\rightarrow\mathbb{R}$ evaluated at $X_i$ and $\epsilon_i\sim \mathcal{N}\left(0, \sigma^2\right)$ models independent observation noise, an additional set of m learnable inducing points $\mathcal{Z}=\{Z_i\}_{i=1}^{m}$, $m\ll n$ are introduced to summarize the information of the entire dataset. These points $Z_i\in\mathcal{Z}$ reside in the same space as the inputs $X_i\in\mathcal{X}$ and are optimized during training to capture the underlying data distribution. In our formulation, $n$ is the total number of instances across all bags, $m$ is the number of learnable prototypical instance representations and we train the SGP layer using batches of instances, where each batch contains instances from the same bag. Let the vector $\mathbf{F}\in\mathbb{R}^n$ contain the values of function $F$ evaluated at each training point $X_i$ and the vector $\mathbf{U}\in\mathbb{R}^m$ be the values of the same function at the points $Z_i\in\mathcal{Z}$. Assuming that the vectors $(\mathbf{F},\mathbf{U})$ are jointly gaussian, the SGP model evaluated at finite sets of inputs $(\mathbf{X},\mathbf{Z})$ can be characterized by the following equations,
\begin{align}
p(\mathbf{F} \mid \mathbf{U}) &= \mathcal{N}\Big(\mathbf{F} \;\Big|\; \boldsymbol{\mu}_\mathbf{X} + \mathbf{\Sigma}_{XZ}\mathbf{\Sigma}_{ZZ}^{-1}(\mathbf{U} - \boldsymbol{\mu}_\mathbf{Z}), \nonumber \\
&\hspace{4.8em} \mathbf{\Sigma}_{XX} - \mathbf{\Sigma}_{XZ}\mathbf{\Sigma}_{ZZ}^{-1}\mathbf{\Sigma}_{ZX} \Big)
\\
p(\mathbf{F}, \mathbf{U}) &= \mathcal{N}\left(
\begin{bmatrix}
\mathbf{F} \\
\mathbf{U}
\end{bmatrix}
\middle\vert
\begin{bmatrix}
\boldsymbol{\mu}_\mathbf{X} \\
\boldsymbol{\mu}_\mathbf{Z}
\end{bmatrix},
\begin{bmatrix}
\mathbf{\Sigma}_{XX} & \mathbf{\Sigma}_{XZ} \\
\mathbf{\Sigma}_{ZX} & \mathbf{\Sigma}_{ZZ}
\end{bmatrix}
\right)
\end{align}
where $\mathbf{\Sigma}_{ZZ}$ is a kernel function evaluating interactions between the inducing points, $\mathbf{\Sigma}_{ZX}$ quantifying correlation between inducing and training points and $\mathbf{\Sigma}_{XX}$ between training points. As will be described in the next section, we will use variational inference to approximate the marginal $p\left(\mathbf{F}\right)=\mathcal{N}\left(\boldsymbol{\mu}_\mathbf{X}, \mathbf{\Sigma}_{\mathbf{XX}}\right)$. For notation relevance, we will denote from now on $\mathbf{F}$ as $\mathbf{A}$ to indicate the attention vector for instances in a WSI.

\subsection{SGPMIL: Sparse Gaussian Process MIL} 
An overview of the SGPMIL pipeline is summarized in~\cref{fig:NNGP_pipeline_figure}. Following the standard attention-based MIL formulation introduced in \cite{ilse2018attention}, the learned function is decomposed into three components: a feature extractor, an attention module and a classifier after a pooling operation. The feature extractor maps each instance to a high-dimensional embedding space, \(f_{\text{e}}: x_{k} \in X_i \mapsto \mathbf{h}_{k} \in \mathcal{H}_i\), where \(\mathcal{H}_i = \{h_{1}, \ldots, h_{K_i}\}\). The attention module is probabilistic: to compute the attention scores, the embeddings are passed through dimensionality-reducing linear layers with appropriate activation functions and then into the SGP layer, which outputs a multivariate normal distribution over the instance-level attention scores. Drawing \(N_s\) Monte Carlo (MC) samples from this learnable distribution produces distinct sets of instance weightings. These weighted embeddings are aggregated via summation, resulting in \(N_s\) bag-level representations. Each bag-level sample is then passed through a classifier \(f_{\text{c}}: \mathbb{R}^d \to \Delta^{C-1}\) to produce an estimate of the conditional distribution \(p(Y \mid X)\). The function decomposition is formally expressed as,

\begin{equation}
    \mathbf{p}_s = f_{\text{c}} \left( \sum_{k=1}^{K_i} A_{s,k} \, \mathbf{h}_k \right)
    \label{eq:mil_decomposition}
\end{equation}
where each $\mathbf{p_s}\in\Delta^{C-1}$ denotes the class probability vector for sample $s$ and $A_{s,k}$ is a matrix where each column contains $s$ samples for each instance $\mathbf{h_k}$ and for which $\frac{1}{S}\sum_{s} A_{s,k}\in\left[0,1\right]$. For $C$ classes, the $(C-1)$-dimensional probability simplex $\Delta^{C-1} = \left\{ \mathbf{p} \in \mathbb{R}^C : p_j \geq 0\ \forall j \in [C],\ \sum_{j=1}^C p_j = 1 \right\}$. 
In turn, the joint probabilistic bag-level model can be written as, 
\begin{equation}
    p\left(\mathbf{Y}, \mathbf{A} , \mathbf{U}\right) = p\left(\mathbf{Y}\mid\mathbf{A},\mathbf{X}\right)p\left(\mathbf{A}\mid\mathbf{U},\mathbf{Z},\mathbf{X}\right)p\left(\mathbf{U}\mid\mathbf{Z}\right)
    \label{eq:full_probabilistic}
\end{equation}
where $\mathbf{A}$ is the attention random variable, $\mathbf{U}$ are the inducing points and $\mathbf{Z}$, $\mathbf{X}$ are the learnable inducing point locations and actual data respectively. Following the work of \cite{schmidt2023probabilistic}, we place a prior over the inducing values $p(\mathbf{U})=\mathcal{N}\left(\boldsymbol{\mu}_{\mathbf{U}}, \Sigma_{\mathbf{ZZ}}\right)$. To approximate the intractable joint distribution $p(\mathbf{A},\mathbf{U})=p(\mathbf{A}\mid \mathbf{U},\mathbf{Z},\mathbf{X})p(\mathbf{U}|\mathbf{Z})$ in \cref{eq:full_probabilistic}, a variational distribution over the inducing values $q(\mathbf{U})=\mathcal{N}\left(\mathbf{U}\mid\mathbf{m}_{\mathbf{U}}, \mathbf{S}_{\mathbf{ZZ}}\right)$ is considered. This leads to a closed form marginal for the attention distribution $q(\mathbf{A}) = \int_{\mathbf{U}} p(\mathbf{A} \mid \mathbf{U}) \, q(\mathbf{U}) \, d\mathbf{U}$ which yields, 

\begin{align}
q(\mathbf{A}) = \mathcal{N}\Big(&\boldsymbol{\mu}_{\mathbf{X}} 
+\Sigma_{\mathbf{XZ}}\Sigma_{\mathbf{ZZ}}^{-1}(\mathbf{m}_{\mathbf{U}} - \boldsymbol{\mu}_{\mathbf{U}}), \nonumber \\
&\Sigma_{\mathbf{XX}} - \Sigma_{\mathbf{XZ}}\Sigma_{\mathbf{ZZ}}^{-1}(\Sigma_{\mathbf{ZZ}} - \mathbf{S}_{\mathbf{ZZ}})\Sigma_{\mathbf{ZZ}}^{-1}\Sigma_{\mathbf{ZX}}\Big)
\label{eq:variational_f}
\end{align}
In our approach, we introduce a learnable linear projection term in the variational posterior mean, defined as \(\boldsymbol{\mu}_{\mathbf{X}} = W_{\mathbf{A}} X + b\). The parameters involved in the optimization include the weights of the projection layers and the classifier, collectively denoted by \(W\); the projection matrix for the real inputs to the SGP \(W_{\mathbf{A}}\); the inducing point set \(\mathbf{Z}\); the kernel hyperparameters \(\theta\) and the variational mean \(\mathbf{m}_{\mathbf{U}}\). The kernel matrix elements are computed using a scaled Radial Basis Function (RBF) kernel function, defined as $\Sigma_{ij} = A \exp\left( - (x_i - x_j)^\top \Theta^{-1} (x_i - x_j) \right) + C$. These components are jointly learned by maximizing the Evidence Lower Bound (ELBO), which is defined as,

\begin{equation}
    \log p(Y) \geq \mathbb{E}_{q(\mathbf{A})} \left[ \log p(Y \mid \mathbf{A}) \right] 
    - \text{KL}\left( q(\mathbf{U}) \,\|\, p(\mathbf{U}) \right),
\end{equation}
Assuming that the data points (WSIs) are i.i.d., the expectation over the variational posterior \(q(\mathbf{A})\) can be decomposed and approximated as
\begin{align}
\mathbb{E}_{q(\mathbf{A})} \left[ \log p(Y \mid \mathbf{A}) \right] 
&= \sum_i \mathbb{E}_{q(\mathbf{A}_i)} \left[ \log p(Y_i \mid \mathbf{A}_i) \right] \nonumber \\
&\simeq \frac{1}{N_s} \sum_{i,s} \log \mathbf{p}_{i,s}
\label{eq:log_likelihood}
\end{align}
where samples drawn from \(q(\mathbf{A})\) are passed through the remainder of the network, producing samples of the class probability vectors \(\mathbf{p}_s\) as defined in \cref{eq:mil_decomposition}. Maximizing \cref{eq:log_likelihood} is equivalent to minimizing a standard Cross-Entropy loss between predicted and observed class probabilities \cite{gal2016dropout}. For inference, the predictive distribution is $p(\mathbf{Y^*}) = \int_{\mathbf{A}} p(\mathbf{Y^*} \mid \mathbf{A}) \, q(\mathbf{A}) \, d\mathbf{A}$ and is approximated by taking MC samples from the learned variational posterior, propagating through the network and again computing the class probability vector as in \cref{eq:mil_decomposition}. To generate MC samples of the attention scores, we apply the local reparameterization trick, rewriting \(\mathbf{A}\) as \(\mathbf{A} = \boldsymbol{m}_{\mathbf{A}} + \sqrt{\text{diag}(\mathbf{S}_{\mathbf{A}})} \odot \epsilon\) with \(\mathbf{A} \sim q(\mathbf{A})\), where \(\boldsymbol{m}_{\mathbf{A}}\) and \(\mathbf{S}_{\mathbf{A}}\) represent the mean and covariance of the variational distribution respectively and \(\epsilon \sim \mathcal{N}(0, I)\) the noise term.

%% file: sections/4_experiments.tex
\subsection{Datasets}
\input{Tables/ablations_table}

\textbf{CAMELYON16}~\cite{CAM16} includes 270 training and 130 test WSIs for breast cancer metastasis detection, with detailed tumor annotations for positive slides. We use 129 test slides due to a corrupted file. WSIs are segmented and patched at $10\times$ magnification~\cite{mammadov2025self_10xcam}, with features extracted via the UNI encoder~\cite{chen2024uni}. We follow~\cite{lu2021clam} for 10-fold cross-validation and report mean bag- and instance-level performance. 

\noindent \textbf{TCGA-NSCLC} For the non-small cell lung carcinoma (NSCLC) subtyping task, we utilize hematoxylin and eosin (H\&E) stained WSIs from The Cancer Genome Atlas (TCGA) to classify lung adenocarcinoma (LUAD) versus lung squamous cell carcinoma (LUSC)~\cite{cooper2018pancancer_tcga,campbell2016distinct_tcga_nsclc}. The current cohort consists of $1{,}006$ slides, with $494$ LUAD and $512$ LUSC cases. WSIs are segmented and patched at $10{\times}$ magnification~\cite{mammadov2025self_10xcam}, with features extracted using UNI~\cite{chen2024uni}. We evaluate performance using 4-fold cross-validation and report mean results across folds.

\noindent \textbf{PANDA}~\cite{panda_dataset} consists of 10,609 WSIs of prostate core needle biopsies, each annotated with an ISUP grade (0–5), resulting in a 6-class classification task. We perform 5-fold cross-validation using stratified splits, where each fold maintains approximately 80/5/15 splits per class for training, validation, and testing respectively. WSIs are segmented into non-overlapping $224 \times 224$ patches at $20\times$ magnification~\cite{panther}, and features are extracted using  UNI~\cite{chen2024uni, panther, zhang20242dmamba}.

\noindent \textbf{BRACS}~\cite{brancati2022bracs} comprises 547 WSIs labeled as Benign, Atypical, or Malignant. We use the predefined train, validation, and test split provided with the dataset, extract non-overlapping $224 \times 224$ patches at $20\times$ magnification~\cite{zhang20242dmamba}, and obtain features using the UNI encoder~\cite{chen2024uni}. All models are trained using 5 random seeds, and we report the mean and standard deviation of each evaluation metric.

\subsection{Implementation Details} 
Our framework, encompassing data preprocessing and model training, is implemented in PyTorch~\cite{Ansel_PyTorch_2_Faster_2024} and PyTorch Lightning~\cite{Falcon_PyTorch_Lightning_2019}. Feature extraction utilizes the TRIDENT toolkit~\cite{trident1, trident2} and the UNI encoder~\cite{chen2024uni}. For the AGP and SGPMIL models, we employ the AdamW optimizer~\cite{loshchilov2019adamw} with a linear warmup followed by a cosine annealing learning rate schedule~\cite{loshchilov2017sgdrstochasticgradientdescent, he2018bagtricksimageclassification}, using a peak learning rate of $6 \times 10^{-4}$ for the TCGA-NSCLC and CAMELYON16 datasets. All other models are trained using their original optimization settings. For the PANDA and BRACS datasets, all models are trained for a maximum of 20 epochs using the AdamW optimizer with cosine annealing and a maximum learning rate of $1 \times 10^{-4}$. AGP and SGPMIL required a slightly higher peak learning rate of $2 \times 10^{-4}$ to ensure convergence within the same number of epochs. The SGP layer is implemented using GPyTorch~\cite{gardner2021gpytorchblackboxmatrixmatrixgaussian}. 

\subsection{Evaluation Metrics} 
For slide-level classification, we report balanced accuracy (ACC), area under the receiver operating characteristic curve (AUC), and adaptive expected calibration error (ACE)~\cite{nixon2019measuring_ACE}, capturing both prediction performance and model calibration. To evaluate statistical significance in all tasks, we conducted a one-sided paired t-test across cross-validation folds, testing whether our method consistently outperforms the baseline.

For patch-level evaluation, we follow the methodology of~\cite{yufei2022bayes}, treating the attention scores produced by each model as instance-level probabilities. The area under the ROC curve (AUC) is computed directly from these continuous scores by comparing them against pixel-level ground truth annotations. For threshold-dependent metrics such as accuracy (ACC), we apply thresholding over a range of values and report the best result per model, following the threshold selection strategy proposed in~\cite{yufei2022bayes}.

%% file: Tables/ablations_table.tex
\begin{table*}[htbp]
\centering
\begin{tabular}{@{\hskip 3pt}l@{\hskip 3pt}r@{\hskip 3pt}c@{\hskip 3pt}c@{\hskip 3pt}c@{\hskip 3pt}c@{\hskip 3pt}c@{\hskip 3pt}c@{\hskip 3pt}c@{\hskip 3pt}c@{\hskip 3pt}c}
\toprule
\multicolumn{5}{c}{} & \multicolumn{3}{c}{WSI} & \multicolumn{2}{c}{INSTANCE} & \multicolumn{1}{c}{Training/Inference}
\\
\cmidrule(lr){6-8} \cmidrule(lr){9-10} \cmidrule(lr){11-11}
\midrule
 & LM & Post & Diag & $N_U$ & ACC & AUC & ACE & AUC & ACE & $t$ (s)
\\
\midrule
\multirow{2}{*}{AGP \cite{schmidt2023probabilistic}}  & \multirow{2}{*}{\ding{55}} & \multirow{2}{*}{$\operatorname{softmax}(\cdot)$} & \multirow{2}{*}{\ding{55}} & $16$ & $.858_{.029}$ &  $.933_{.029}$ & $.089_{.007}$ & $.946$ & $.091$ & $31.0$/$6.0$ 
\\

& & & & $80$ & $.883_{.021}$ & $.954_{.019}$ & $.069_{.013}$ & $.953$ & $.080$ & $31.0$/$6.0$
\\

& \ding{55} & $\sigma(\cdot)$ & \ding{55} & $80$ & $.936_{.015}$ & $.970_{.009}$ & $.049_{.014}$ & $.822$ & $.320$ & $31.0$/$6.0$
\\

& \ding{51} & $\operatorname{softmax}(\cdot)$ & \ding{55} & $80$ & $.971_{.006}$ &  $.976_{.007}$  & $.027_{.008}$ & $.821$ & $.173$ & $31.0$/$6.0$
\\

& \ding{51} & $\sigma(\cdot)$ & \ding{55} & $80$ & $.972_{.014}$ &  $.979_{.016}$  & $.065_{.045}$ & $\underline{.970}$ & $.103$ & $31.0$/$6.0$ 
\\

& \ding{51} & $\operatorname{softmax}(\cdot)$ & \ding{51} & $80$ & $\underline{.973_{.007}}$ &  $\underline{.980_{.009}}$  & $\underline{.024_{.007}}$ & $.965$ & $\underline{.070}$
 & $\mathbf{9.0}$/$\mathbf{1.0}$
\\

\cmidrule(lr){2-11}
\multirow{2}{*}{SGPMIL} & \multirow{2}{*}{\ding{51}} & \multirow{2}{*}{$\sigma(\cdot)$} & \multirow{2}{*}{\ding{51}} & $16$ & $.971_{.016}$ &  $.977_{.020}$ & $.045_{.022}$ & $.963$ & $.271$ &   $\mathbf{9.0}$/$\mathbf{1.0}$ 
\\

& & & & $80$ & $\mathbf{.980_{.007}}$ &  $\mathbf{.986_{.005}}$ & $\mathbf{.021_{.005}}$ & $\mathbf{.973}$ & $\mathbf{.051}$ & $\mathbf{9.0}$/$\mathbf{1.0}$   
\\

\bottomrule
\end{tabular}
\caption{Ablation study on the CAMELYON16 dataset (10-fold cross-validation) evaluating architectural differences to vanilla AGP model. Variants differ in mean parameterization (LM), post-SGP activation (Softmax/Sigmoid), use of covariance diagonal (Diag) and number of inducing points ($N_U$). Metrics include WSI/Instance-level ACC, AUC, and average training/inference time $t$~(s).}
\label{tab:agp_nngp_ablations}
\end{table*}

%% file: sections/5_results.tex
\subsection{Ablations}
We showcase the performance gains achieved by the modifications introduced in our framework, relative to a baseline implementation of sparse GPs for MIL~\cite{schmidt2023probabilistic}. \Cref{tab:agp_nngp_ablations} presents the improvement in bag-level performance our contributions have against the vanilla methodology. The ablations systematically vary four components: attention normalization (implemented via softmax or sigmoid), the mean parameterization in the variational posterior (LM), the use of diagonal covariance (Diag), and the number of inducing points ($N_\mathbf{U}$). Analysis on the effect of varying the number of inducing points is provided in supplementary Table~11.

The first improvement over the vanilla AGP model is achieved by introducing relaxed attention normalization, replacing the softmax constraint $\sum_k A_{s,k} = 1$ with a sigmoid-based expectation constraint $\frac{1}{S} \sum_s A_{s,k} \in [0,1]$. This relaxation allows the model to naturally assign high attention to multiple instances with task-relevant characteristics, yielding performance gains in both bag- and instance-level.
Next, we observe a substantial improvement after introducing a feature-scaling term in the variational mean $\boldsymbol{\mu_X}=\mathbf{W_A}X+\mathbf{b}$, which replaces the constant $\mathbf{\mu_X}=C$ used in AGP. This adjustment improves expressiveness and makes optimization over kernel hyperparameters and inducing point locations $\mathbf{Z}$ more effective.
The combination of relaxed normalization and feature scaling leads to further gains, especially in instance-level accuracy and calibration.

Finally, we address the numerical instability and speed bottlenecks of AGP by replacing the full covariance matrix with its diagonal approximation, eliminating the need for Cholesky decomposition during training. This leads to much faster and more stable training and inference, without sacrificing predictive performance. Comprehensive training and inference time comparisons against all baselines are provided in supplementary Table~6. To further underscore the stability gains of our modifications, we compute the Lipschitz constant as a surrogate of stability of each variant $L(f)\;\le\;\prod_{\ell=1}^{L}\,\sigma_{\max}(W_\ell)\, \prod_{\ell=1}^{L-1} L_{\sigma_\ell}$,  where $\sigma_{max}$ denotes the largest singular value, and $L_{\sigma_l}$ the Lipschitz value of activation functions, obtaining $L_{\mathrm{AGP}}=105.5_{25.9}$ vs. $L_{\mathrm{SGPMIL}}=38.1_{9.1}$, a major reduction in input sensitivity, underscoring the significance of our contributions.
Overall, introducing feature scaling in the variational mean, diagonal covariance approximation, and relaxed attention normalization leads to improved bag-level performance, faster convergence, and more focused instance-level attention.

\begin{figure*}
   \centering
   \includegraphics[width=0.95\linewidth]{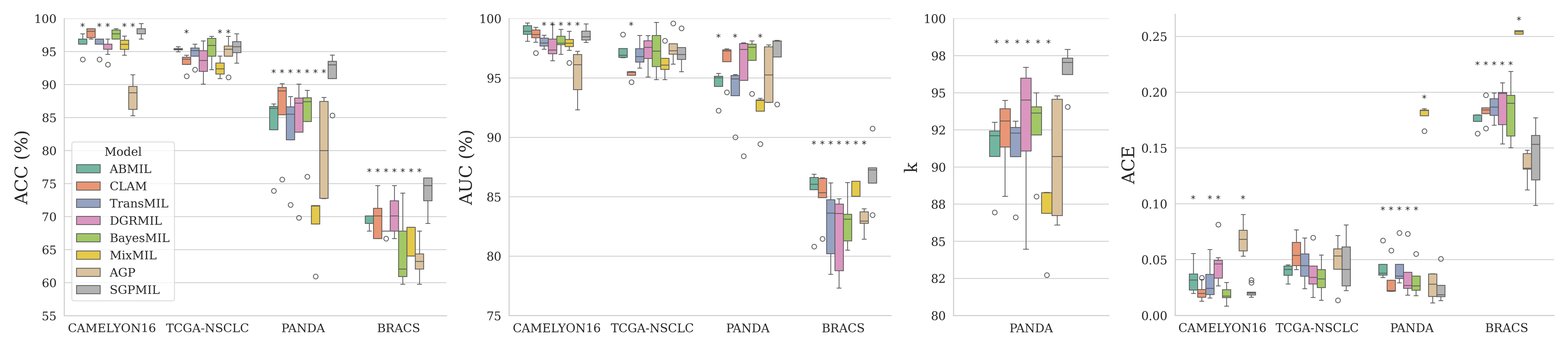}
   \caption{Slide-level performance across multiple datasets. ${}^\star$ denotes statistical significance ($p < 0.05$) based on one-sided paired t-tests.}
   \label{fig:slide-level_boxplots}
\end{figure*}

\subsection{Bag-Level Performance}

\Cref{fig:slide-level_boxplots} reports the performance of SGPMIL and several baselines across the different histopathology datasets and tasks together with one-sided paired t-tests to evaluate significance for all datasets. 

On CAMELYON16, SGPMIL achieves significantly higher accuracy than all models except CLAM and BayesMIL, and significantly higher AUC than all models except ABMIL and CLAM. In terms of calibration, BayesMIL, CLAM, and SGPMIL perform comparably well, with SGPMIL achieving statistically significantly lower ACE than the remaining approaches, while BayesMIL attains the lowest ACE overall, consistent with its probabilistic formulation. For TCGA-NSCLC, performance is uniformly high across models; however, SGPMIL is significantly better than CLAM in ACC and AUC, outperforms MixMIL and AGP in ACC and achieves better calibration than MixMIL.

The PANDA dataset presents a greater challenge due to its multi-class nature. Here, our method ranks first with statistically significant improvements in both ACC and $\kappa$---the preferred metric over AUC for this task---outperforming the next-best method by margins of $3.3\%$, and $2.2\%$ respectively, while achieving comparable calibration to AGP and superior calibration to the other baselines. A similar trend is observed in BRACS, where our approach leads in both ACC and AUC by $3.3\%$ and $1.5\%$ in a significant manner, again performing similarly to AGP in calibration and better than other baselines. We note that DGRMIL was not originally evaluated in multi-class settings. To integrate it, we substitute the binary cross-entropy loss with a cross-entropy loss followed by a softmax activation, updating the negative center when the bag belongs to the 'normal' class and the positive center in any other case.
These results highlight the generalization capabilities of the SGP-based formulation, particularly in complex multi-class scenarios with heterogeneous tissue morphologies.

\subsection{Instance-Level Localization}

\input{Tables/instance_level_combined}

\begin{figure*}
    \centering
    \includegraphics[width=0.78\linewidth]{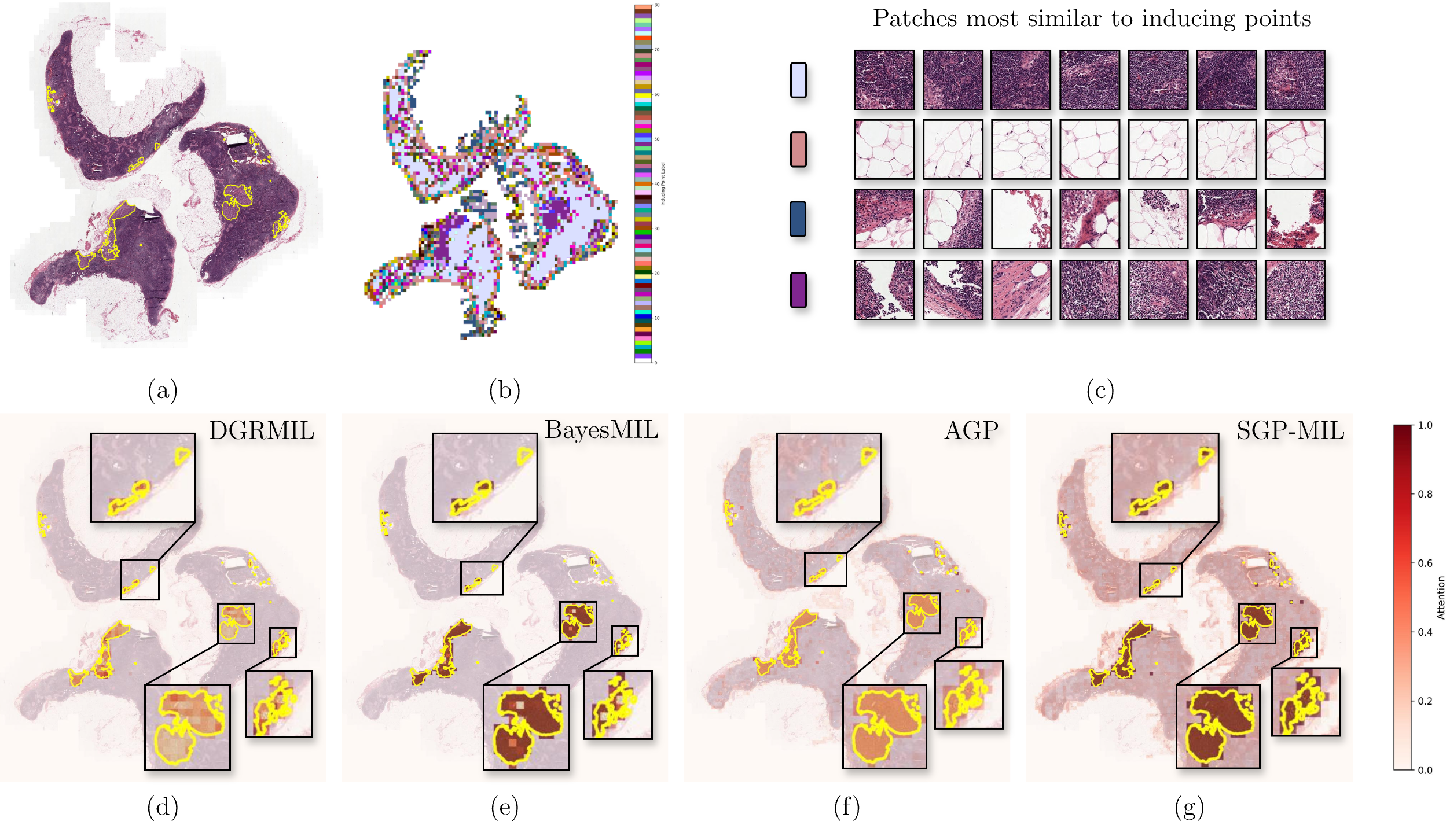}
\caption{Inducing-point structure and normalized attention heatmaps on a CAMELYON16 test slide.
Top row: (a) slide with ground-truth annotations; (b) inducing-point label map; (c) top-7 most similar patches for the most representative inducing points.
Bottom row: normalized attention heatmaps (scores in $[0,1]$). For DGRMIL, we use the cls-token attention; for BayesMIL, AGP, and SGPMIL, we plot the mean attention per patch. Yellow contours denote ground-truth annotations. See Supplementary Figure~3 for additional visualizations.}
    \label{fig:heatmaps}
\end{figure*}

A key strength of our approach lies in its ability to identify relevant instances within each bag. We interpret the attention score as a proxy for the model's capability to focus on patches with task-specific characteristics. For each dataset, we report the performance across all folds. For ABMIL and CLAM, attention scores are extracted from the gated attention mechanism. In TransMIL, we perform mean fusion of the attention matrices of attention heads within each of the two attention blocks and apply attention roll-out \cite{abnar2020quantifying_attentionrollout} to compute final scores. DGRMIL follows the method described by \cite{zhu2024dgrmil}, using the global token attention. For BayesMIL and MixMIL, we use the mean of the attention variational posterior per patch. For AGP and SGPMIL, we derive attention scores by averaging post-activation values---softmax for AGP and sigmoid for SGPMIL---across samples drawn from the learned SGP variational posteriors. All attention scores are subsequently normalized to the range $a \in [0, 1]$ and evaluated against ground-truth annotations on test-set WSIs. In \Cref{tab:instance_level_combined}, SGPMIL consistently outperforms all baselines by over 4\% in ACC and 2\% in AUC in a statistically significant manner; calibration-wise AGP and SGPMIL are more calibrated than all other baselines in CAMELYON16 while on BRACS, SGPMIL achieves the best calibration among all methods with non-trivial performance, as ABMIL, CLAM and MixMIL perform at chance level. Moreover, SGPMIL provides focused attention on diagnostically salient regions.

\subsection{Inducing Points Capture Patch Variability}

Recent MIL work has highlighted the importance of bag-level representations, noting that noisy instance features can negatively impact classification performance~\cite{zhu2025effective-MIL_dropout}. In our approach, the learnable inducing points act as task-adaptive prototypes that capture salient morphological patterns and core structural features across tissue slides, shown in \cref{fig:heatmaps}. 

In \Cref{fig:heatmaps}(a), we visualize the slide-level ground truth annotations. In (b), we show a label map where each patch embedding is assigned the label of its most similar inducing point (based on cosine similarity). This visualization reveals that distinct inducing points specialize in capturing different tissue types—e.g., tumor regions, dense stromal tissue, and interface zones such as tumor-stroma boundaries. Finally, in \Cref{fig:heatmaps}(c), we display the top-7 most similar patches for the inducing points that match the largest number of patches. The visual coherence across patches associated with each inducing point demonstrates that the model learns meaningful and morphologically consistent prototypes. Additional inducing point label maps and HoverNet-based cancer/nucleus-type distributions~\cite{graham2019hover} are shown in Supplementary Figures~4, 5 and 6.

\subsection{Prediction Uncertainty Estimation}
\input{Figures/uncertainty}

In \Cref{fig:prediction_probability_uncertainty}, we visualize the prediction uncertainty for correctly and incorrectly classified WSIs. We compute the standard deviation of the predicted class probabilities across samples and compare the resulting distributions using a two-sample Welch’s $t$-test~\cite{welch1947generalization_ttest}. We consider BRACS and PANDA datasets as CAMELYON16 and TCGA-NSCLC contain too few misclassified slides for reliable variance estimation (due to high ACC and AUC).
A statistically significant difference ($p < 0.05$) in prediction uncertainty is observed between correct and incorrect predictions across datasets. This indicates that higher uncertainty is associated with misclassified slides, suggesting that our model’s uncertainty estimates are well-calibrated—an essential property for deployment in safety-critical clinical workflows.

%% file: Tables/instance_level_combined.tex
\begin{table}[ht]
    \centering
    \resizebox{\linewidth}{!}{%
    \begin{tabular}{l@{\hskip 2pt}c@{\hskip 2pt}c@{\hskip 2pt}c@{\hskip 2pt}c@{\hskip 2pt}c@{\hskip 2pt}c}
    \toprule
    & \multicolumn{3}{c}{CAMELYON16} & \multicolumn{3}{c}{BRACS} \\
    \cmidrule(lr){2-4} \cmidrule(lr){5-7}
    & ACC & AUC & ACE & ACC & AUC & ACE \\
    \midrule
    ABMIL~\cite{ilse2018attention} & $\ .775^{\star}\ $ & $\ .910^{\star}\ $ & $\ .191^{\star}\ $ & $\ .514^{\star}\ $ & $\ .560^{\star}\ $ & $\ .062\ $ \\
    CLAM~\cite{lu2021clam} & $\ .678^{\star}\ $ & $\ .792^{\star}\ $ & $\ .245^{\star}\ $ & $\ .522^{\star}\ $ & $\ .590^{\star}\ $ & $\ .061\ $ \\
    TransMIL~\cite{shao2021transmil} & $\ .578^{\star}\ $ & $\ .803^{\star}\ $ & $\ .110^{\star}\ $ & $\ .696^{\star}\ $ & $\ \underline{.852}^{\star}\ $ & $\ .188^{\star}\ $ \\
    DGRMIL~\cite{zhu2024dgrmil} & $\ .710^{\star}\ $ & $\ .820^{\star}\ $ & $\ .264^{\star}\ $ & $\ .673^{\star}\ $ & $\ .720^{\star}\ $ & $\ .207^{\star}\ $ \\
    BayesMIL~\cite{yufei2022bayes} & $\ .787^{\star}\ $ & $\ .855^{\star}\ $ & $\ .166^{\star}\ $ & $\ .568^{\star}\ $ & $\ .684^{\star}\ $ & $\ .096^{\star}\ $ \\
    MixMIL~\cite{mixmil} & $\ .563^{\star}\ $ & $\ .670^{\star}\ $ & $\ .256^{\star}\ $ & $\ .509^{\star}\ $ & $\ .528^{\star}\ $ & $\ \mathbf{.033}\ $ \\
    AGP~\cite{schmidt2023probabilistic} & $\ \underline{.864}^{\star}\ $ & $\ \underline{.950}^{\star}\ $ & $\ \mathbf{.046}\ $ & $\ \underline{.718}^{\star}\ $ & $\ .832^{\star}\ $ & $\ .188^{\star}\ $ \\
    \textbf{SGPMIL} & $\ \mathbf{.910}\ $ & $\ \mathbf{.979}\ $ & $\ \underline{.059}\ $ & $\ \mathbf{.765}\ $ & $\ \mathbf{.899}\ $ & $\ \underline{.050}\ $ \\
    \bottomrule
    \end{tabular}}
    \caption{Instance-level performance on CAMELYON16 and BRACS for each model. Metrics shown: accuracy (ACC), AUC and ACE. Bold denotes best result. ${}^\star$ denotes statistical significance ($p < 0.05$) based on one-sided paired t-tests.}
    \label{tab:instance_level_combined}
\end{table}

%% file: Figures/uncertainty.tex
\begin{figure}[h!]
    \centering
    \includegraphics[width=0.82\linewidth]{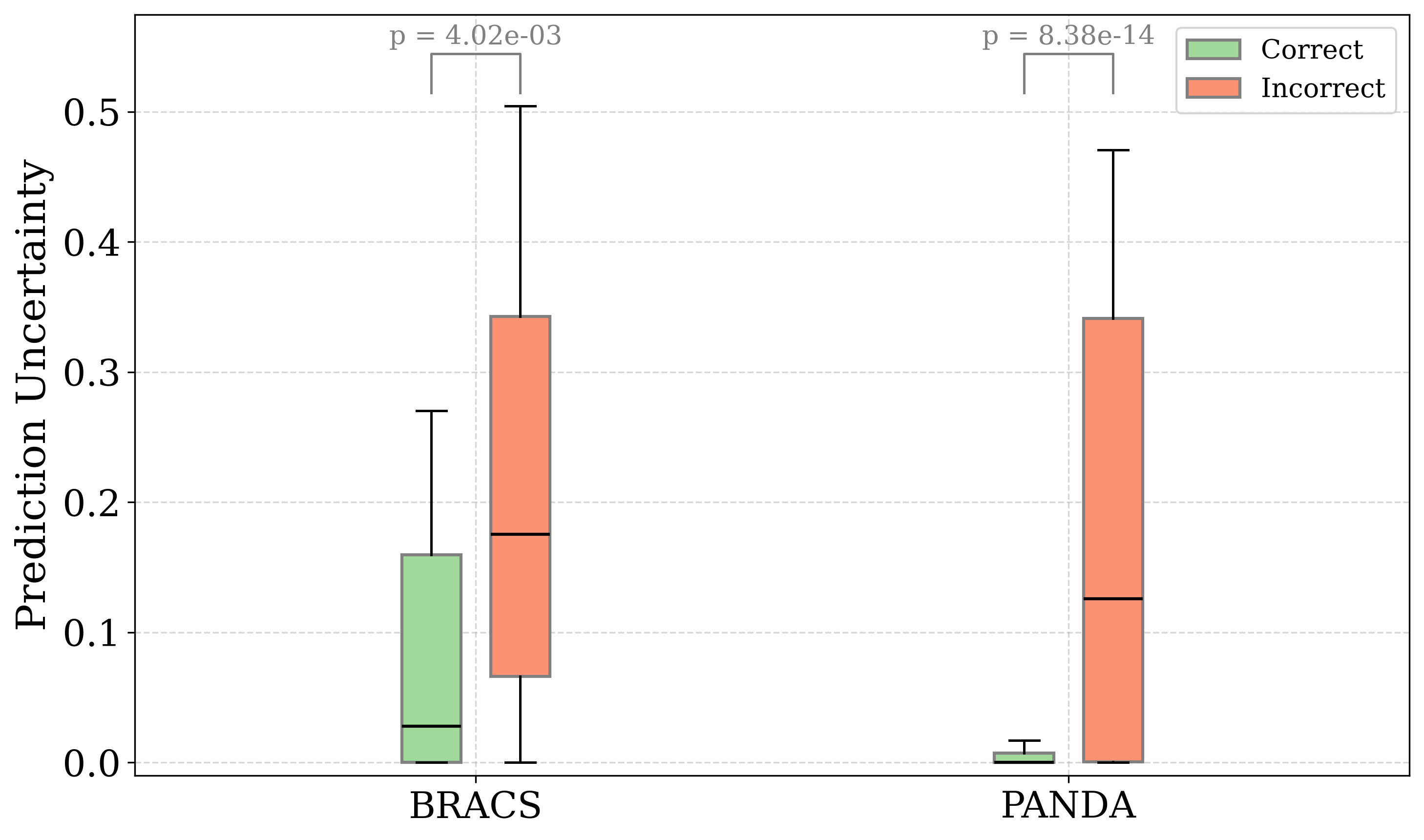}
    \caption{Prediction uncertainty for correctly (green) and incorrectly (red) classified WSIs in the BRACS (left) and PANDA (right) datasets. Each boxplot pair shows the distribution of standard deviation of predicted class probabilities. A statistically significant difference in uncertainty is observed in both datasets ($p < 0.05$, Welch’s $t$-test~\cite{welch1947generalization_ttest}).}
    \label{fig:prediction_probability_uncertainty}
\end{figure}

%% file: sections/6_conclusions.tex
We introduced SGPMIL, a robust and scalable probabilistic attention-based MIL framework built on SGP, designed for improved instance- and bag-level prediction in histopathology tasks. By learning a variational posterior over attention scores and leveraging a flexible kernel, SGPMIL captures semantic relationships among instances while enabling principled uncertainty quantification. Our method achieves leading instance-level performance, competitive bag-level accuracy, and addresses key limitations of prior probabilistic MIL approaches, particularly in terms of numerical stability and scalability. A limitation of this work is that we focus exclusively on unimodal visual inputs. Extending SGPMIL to multimodal MIL settings—such as integrating clinical or genomic metadata alongside WSIs—remains an exciting direction for future research.

%% file: sections/8_aknowledgements.tex
\paragraph{Acknowledgments} This work has been partially supported by project MIS 5154714 of the National Recovery
and Resilience Plan Greece 2.0 funded by the European Union under the NextGenerationEU Program.
Hardware resources were granted with the support of GRNET. 
We also gratefully acknowledge the DATAIA program for supporting JD as a visiting professor at Université Paris-Saclay.